# Beyond the Lungs: Extending the Field of View in Chest CT with Latent Diffusion Models

Lianrui Zuo[a], Kaiwen Xu[b], Dingjie Su[c], Xin Yu[c], Aravind R. Krishnan[a], Yihao Liu[a], Shunxing Bao[a], Thomas Li[d], Kim L. Sandler[e], Fabien Maldonado[f,g], and Bennett A. Landman[a,c,d]

[a]Department of Electrical and Computer Engineering, Vanderbilt University, Nashville, United States
[b]Insitro, South San Francisco, United States
[c]Department of Computer Science, Vanderbilt University, Nashville, United States
[d]Department of Biomedical Engineering, Vanderbilt University, Nashville, Unites States
[e]Department of Radiology, Vanderbilt University Medical Center, Nashville, Unites States
[f]Department of Medicine, Vanderbilt University Medical Center, Nashville, United States
[g]Department of Thoracic Surgery, Vanderbilt University Medical Center, Nashville, United States

## ABSTRACT

The interconnection between the human lungs and other organs, such as the liver and kidneys, is crucial for understanding the underlying risks and effects of lung diseases and improving patient care. However, most research chest CT imaging is focused solely on the lungs due to considerations of cost and radiation dose. This restricted field of view (FOV) in the acquired images poses challenges to comprehensive analysis and hinders the ability to gain insights into the impact of lung diseases on other organs. To address this, we propose SCOPE (Spatial Coverage Optimization with Prior Encoding), a novel approach to capture the inter-organ relationships from CT images and extend the FOV of chest CT images. Our approach first trains a variational autoencoder (VAE) to encode 2D axial CT slices individually, then stacks the latent representations of the VAE to form a 3D context for training a latent diffusion model. Once trained, our approach extends the FOV of CT images in the z-direction by generating new axial slices in a zero-shot manner. We evaluated our approach on the National Lung Screening Trial (NLST) dataset, and results suggest that it effectively extends the FOV to include the liver and kidneys, which are not completely covered in the original NLST data acquisition. Quantitative results on a held-out whole-body dataset demonstrate that the generated slices exhibit high fidelity with acquired data, achieving an SSIM of 0.81.

## 1. INTRODUCTION

The impact of human lung diseases often extends beyond the respiratory system, affecting other organs such as the liver and the kidneys. For instance, chronic obstructive pulmonary disease (COPD) can lead to hepatic congestion,[1] while pulmonary complications are common in patients with chronic kidney disease (CKD).[2,3] Extensive research has been conducted to study the interconnection between the human lung and other organs, emphasizing the importance of viewing the human body as an integrated system. For example, Kagohashi et al. explored association between clinicopathological features and liver metastases from lung cancer.[4] Goldberg et al. investigated the interplay between lung and heart diseases in the context of liver conditions.[5] Understanding these interconnections is crucial for comprehensive patient care, treatment planning, and monitoring disease progression.

Despite the significance of these inter-organ relationships, most existing lung screening protocols focus solely on the lung region, primarily due to considerations of cost and radiation dose. Additionally, incidentalomas are frequent and generate unnecessary downstream procedures with complications, increasing cost and morbidity. For example, the National Lung Screening Trial (NLST)[6] recommends a computed tomography (CT) scanning protocol that exclusively covers the lung region. As shown in Fig. 1**(a)**, an improved body part regression (BPR) algorithm[7] was applied to randomly selected CT

---

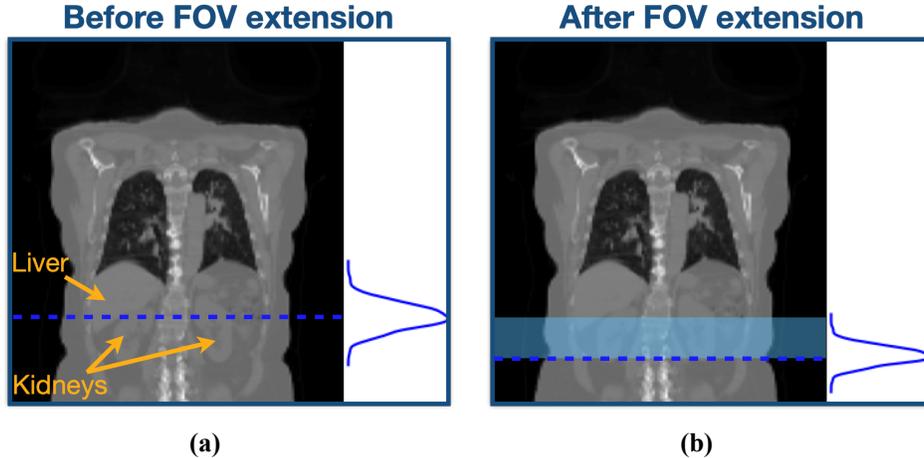

|  (a)  |  (b)  |

Figure 1. Estimated distribution of FOV (blue curve) of the NLST dataset overlaid on top of a reference CT image. **(a)** While lung region is covered in all NLST data, liver and kidney regions are only partially covered in the NLST. **(b)** After FOV extension using SCOPE, we extend FOV of the NLST data beyond the lung, covering the liver and kidney regions, as highlighted in shaded blue area.

images ($N = 100$) from the NLST to estimate the overall field of view (FOV) of the acquisition. The distribution of FOV over the sampled dataset is shown on the right (blue curve). These data reveal that most images primarily cover the lung region, with limited coverage of the liver and kidney regions. This restricted FOV poses significant challenges for systematically studying the interplay between the lungs and other organs. Consequently, valuable information that could enhance our understanding of multi-organ impacts of lung diseases may be overlooked.

Considering that the underlying anatomical relationships between the lungs and other organs, such as the liver and kidneys, are inherently present in the CT images, we ask an intriguing question: Can we train a model to study these relationships and recover them when the dataset is partial? Specifically, if a CT acquisition covers the lungs and only partially includes the livers and kidneys, the model could infer the missing information based on the acquired data and the inherent relationships it learned during training. This inference would extend the FOV in a completely data-driven way, thus enabling a more comprehensive analysis. There have been existing studies that explore FOV extension based on similar principles. For instance, Xu et al.[8,9] extended the FOV in the axial plane to fill in the missing subcutaneous fat due to truncation. The study demonstrated consistent body composition analysis and showed strong association between lung cancer risk and body composition after FOV extension. In a different but related vein, super-resolution techniques[10] leverage prior knowledge learned during training to infer high-frequency information from low-resolution inputs, even in the presence of slice gaps in medical images. Furthermore, image-to-image translation tasks, such as image harmonization,[11] imputation,[12,22] and inpainting,[13] also apply prior knowledge to fill in missing data during inference time. These related works highlight the potential of using learned relationships to recover comprehensive information from incomplete datasets.

Inspired by the existing works,[8,9] we propose Spatial Coverage Optimization with Prior Encoding (SCOPE) to address the limited FOV issue in chest CT imaging and to enable studying inter-organ relationships on more datasets. It is worth mentioning that our approach is the first to investigate extending the FOV in the z-direction, which presents additional challenges due to the need to generate new axial slices while maintaining anatomical coherence. SCOPE employs a variational autoencoder (VAE)[14] to encode 2D axial CT slices, which are then used to form a 3D context for training a latent diffusion model (LDM). SCOPE generates new axial slices in a zero-shot manner; it captures the inherent inter-organ relationships as prior knowledge and directly applies this prior knowledge during application. To summarize, the innovations of the paper are twofold. First, we propose SCOPE, a novel method to extend the FOV of chest CT images without the need of training on complete FOV dataset, such as whole-body CT. Second, we use an LDM to transform the high-dimensional 3D nature of CT images into a 2D problem, which reduces computation burden while maintaining 3D context information to the human anatomy.

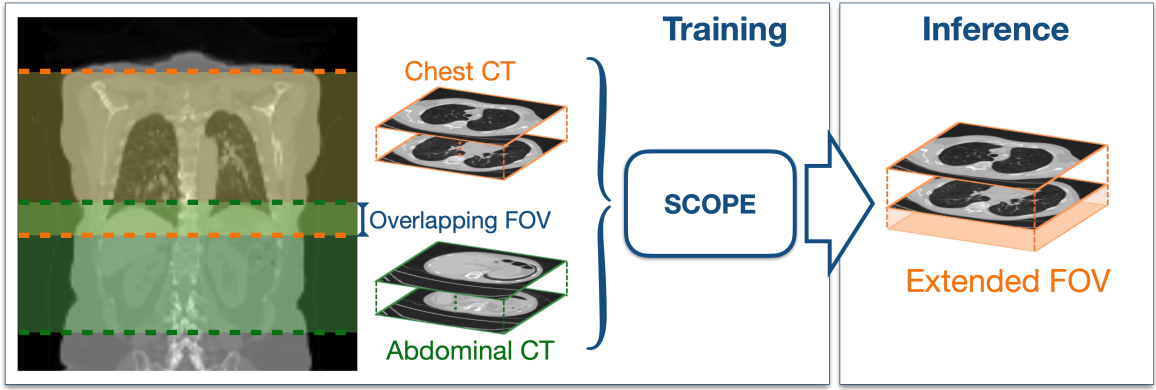

Figure 2. Illustration of the intuition behind SCOPE. The figure shows two example datasets: a chest CT dataset (as highlighted in orange) and an abdominal CT dataset (as highlighted in green) overlaid on a template whole-body CT. Despite focusing on different regions, the two datasets have overlapping FOVs, enabling the pooling of these datasets to train SCOPE. During inference, SCOPE leverages the learned inter-organ relationships to extend the FOV, incorporating regions beyond the initial chest scans.

## 2. METHODS

### 2.1 Bridging the Chest and Abdomen

The core idea behind SCOPE is to feed the model with data that cover both the lungs and the liver and kidneys during training, enabling it to capture the inherent anatomical relationships between these organs. Ideally, this would involve training on large FOV datasets, such as whole-body CT images. However, such comprehensive datasets are not always available, and even when they are, their quantity is often limited.

To circumvent this issue, we propose training SCOPE on two partial datasets: chest CT and abdominal CT. Despite each dataset focusing primarily on its respective organs with limited FOVs, they typically have overlapping regions, as highlighted in Fig. 2. The overlapping regions between the two datasets serve as a "bridge" that allows this inter-organ relationship to be captured across the lungs, liver, and the kidneys during training. Once trained, the model can extend the FOV during inference time, even when only partial data are available.

### 2.2 Framework

#### 2.2.1 Data Preprocessing

Our chest and abdominal CT datasets both have image dimensions of 512 × 512 × $S$, where $S$ is the number of axial slices, and it generally varies from image to image. Each CT image has slice thickness of 3mm without a slice gap. Before training SCOPE, we downsampled the image in axial plane into 256×256×$S$ with a Gaussian blurring as an anti-aliasing filter. The downsampling is to reduce computational consumption while maintaining anatomical information. All image intensities were then clipped to [−1024,3072] Hounsfield Units and normalized to have range of [−1,1]. Finally, we randomly select segments of consecutive $N_s$ slices per 3D image, resulting in a final dimension of 256 × 256 × $N_s$ per data sample. The implication of $N_s$ is discussed in Sec. 2.2.2.

#### 2.2.2 Encoding Prior Knowledge with LDM

As shown in Fig. 3, SCOPE has a two-stage training strategy as in most LDMs.[15] During training, each 2D axial slice from the combined chest and abdominal dataset is first encoded using a VAE. The VAE maps each slice into a 4096-dimensional latent space, compressing the high-dimensional CT images while preserving the essential features. The training objective in Stage #1 is to reduce the VAE loss, which is defined as

$$L_{\text{VAE}} = \|\hat{x} - x\|_1 + \lambda \mathcal{D}_{KL} \left[ p(z|x) \,\|\, p(z) \right], \qquad (1)$$

where $D_{KL}$ is Kullback-Leibler divergence, and $p(z)$ is standard Gasussian distribution $\mathcal{N}(0,I)$. $\lambda$ is a hyperparameter. It is worth noting that we choose a VAE rather than an autoencoder (AE) because VAE allows a smoother and well-defined latent space. This helps the decoder to better appreciate the latent representations generated by the LDM during inference time.

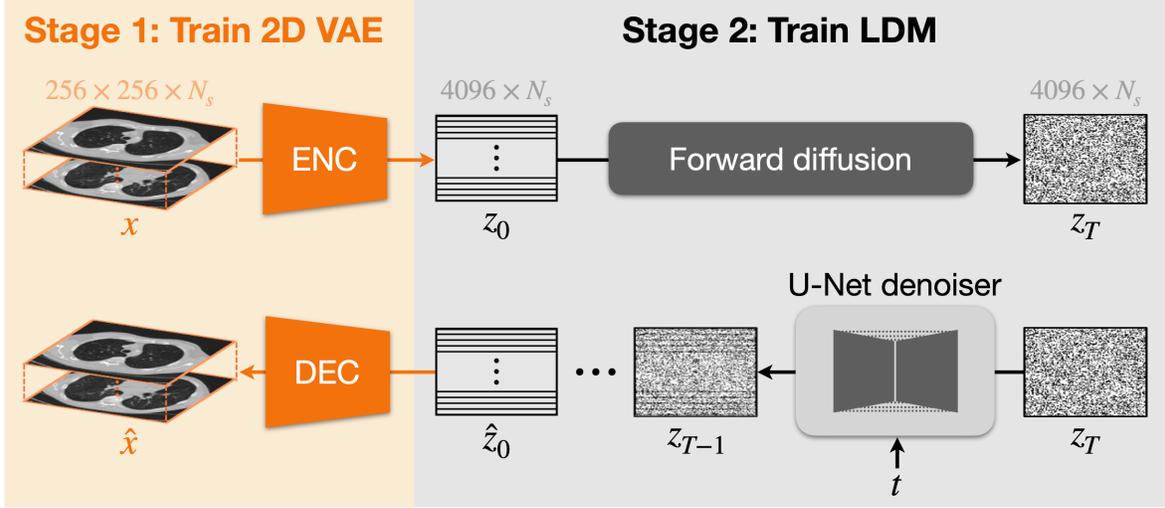

Figure 3. SCOPE has a two-stage training strategy as in most LDMs. First, a variational autoencoder maps an image x into a latent representation z. z's are then stacked to form a 3D context before training a diffusion model. $N_s$ = 64 in our implementation.

After encoding all $N_s$ slices, we stack the latent representations to form a 3D context with dimension of $4096 \times N_s$. This stacked representation allows us to train an LDM[15–17] that captures the spatial relationships across all $N_s$ slices with reduced computational burden. The forward diffusion process is defined as

$$q(z_t|z_{t-1}) = \mathcal{N}(z_t; \sqrt{1-\beta_t}z_t - 1, \beta_t I) \quad (2)$$

where $\beta_t$ controls the level of noise being injected at each step $t$ during the forward diffusion process. The reverse process is parameterized by a neural network denoiser $\epsilon_\theta(z_t;t)$, which is trained to predict the noise added at each step $t$. The training objective is defined as minimizing the following loss function:

$$L_{\text{LDM}} = \|\epsilon - \epsilon_\theta(z_t;t)\|_2^2 \quad (3)$$

where $\epsilon \sim \mathcal{N}(0,I)$ is the Gaussian noise, and $t$ is a randomly sampled time step between 0 and $T$. By training the LDM on the stacked latent representations, SCOPE learns to model the complex anatomical structures and relationships in the latent space $z$, capturing the prior anatomical knowledge of the human body. During inference, this prior knowledge allows SCOPE to infer missing information based on the latent representations of available slices and the learned anatomical relationships.

### 2.2.3 Zero-shot FOV Extension

After training the LDM, SCOPE achieves zero-shot FOV extension without the need for simulating missing slices during training. The core idea is inspired by the RePaint method.[13] The design of SCOPE transforms the problem of synthesizing new slices to cover extended FOV into synthesizing new rows of the latent representation $z$ from the available $z$'s. Specifically, we modify the estimated latent representation $\hat{z}_t$ at each time step $t$ by incorporating the real $\hat{z}_t$ values (with the corresponding level of noise) from the acquired data, i.e.,

$$\hat{z}_t \leftarrow z_t \circ \mathcal{M} + \hat{z}_t \circ (1 - \mathcal{M}) \quad (4)$$

where $\mathcal{M}$ has the same dimension as $z$, and it takes the values of 1 for positions corresponding to acquired slices and 0 for positions corresponding to unavailable slices. The operator "∘" indicates element-wise multiplication. For slices that need to be predicted, the reverse diffusion process estimates the corresponding missing $z$ values, and these $z$ values are then sent to the VAE decoder to generate new slices. The $z$ values from the acquired data guide the diffusion to fill in the missing pieces based on the learned prior anatomical knowledge. It is worth reemphasizing that Eq. 4 is used solely during inference, making SCOPE completely zero-shot.

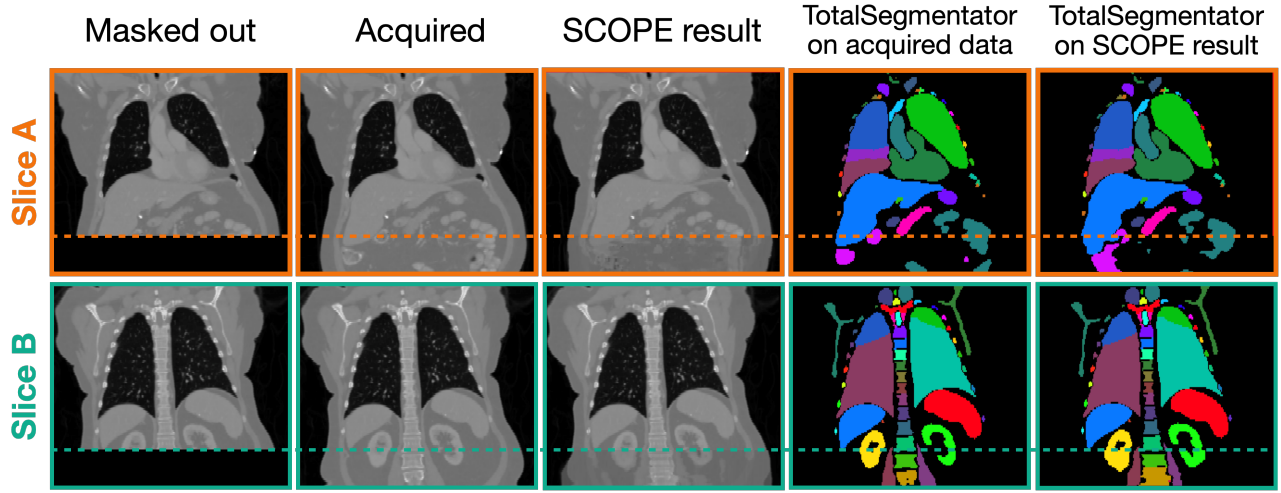

Figure 4. We masked out the lower abdominal region of full FOV CT images to simulate the limited FOV of the NLST dataset. SCOPE was then applied to extend the FOV on the masked-out data. The right two columns show segmentation results of TotalSegmentator[20] on the acquired image and SCOPE-extended image, respectively.

### 2.3 Implementation details

In our implementation, the number of stacked slices $N_s$ is chosen to be 64. Since our training data have slice thickness of 3mm, $N_s$ = 64 covers approximately 20cm of the human body, providing 3D context for the LDM to capture. For the VAE, we utilized the implementation provided in MONAI 1.2.[18] The denoiser network $\epsilon_\theta(z_t;t)$ in our LDM is a U-Net[19] with four downsampling levels. We employed the cosine noise scheduling as recommended in the literature.[16] The number of diffusion steps is set to 1000. For training the VAE and LDM, we used data from $N$ = 500 subjects from the NLST chest dataset (no lung cancer cohort) and an additional 300 subjects from an in-house abdominal dataset. All models were trained on a NVIDIA A6000 GPU with 48 GB of RAM.

## 3. EXPERIMENTS AND RESULTS

### 3.1 Evaluation on held-out body CT data

To evaluate the performance of SCOPE, we conducted both qualitative and quantitative experiments using a held-out dataset of body CT images ($N$ = 10) that cover both the chest and abdominal regions. The body CT images were preprocessed following the same procedure as described in Sec. 2.2.1. After preprocessing, we then masked out the lower abdominal regions of the images to simulate the limited FOV of the NLST dataset. As shown in Fig. 4 left, the masked-out image has a similar FOV as the NLST data, which only partially cover the liver and the kidneys (NLST FOV shown in Fig. 1**(a)**). SCOPE was then employed to extend the FOV by generating 30 additional axial slices (equivalent to 9cm). The synthetic axial slices are concatenated with the acquired slices to generate a 3D volume, as shown in Fig. 4 from coronal view. To further study the anatomical fidelity of the synthetic images, we ran TotalSegmentator[20] on both the acquired ground truth image and the synthetic image. Figure 4 right shows that SCOPE not only generates realistic-looking images, but also produces segmentation results that show strong agreement with the acquired images. A notable property of SCOPE is that the information in the acquired slices does not change during imputation. This is due to the 2D VAE design of SCOPE, which allows each slice to be processed individually while maintaining the 3D context in latent space.

To quantitatively evaluate SCOPE, we calculated the structural similarity index (SSIM)[21] and peak signal-to-noise ratio (PSNR) given different number of imputed slices. Different number of imputed slices impacts the number of slices that can be used as conditions in Eq. 4, thus impacting the overall performance of SCOPE. As shown in Table 1, SCOPE has the best performance when the number of missing slices is 10 (equivalent to 3cm of the human body to be imputed). As the number of imputed slices becomes larger, SCOPE has decreased performance, because it has less slices to use as condition.

Table 1. Performance metrics for SCOPE with respect to the number of predicted slices. Slice thickness is 3mm. Numbers are reported in "Mean"±"Standard deviation".

| # of imputed slices | 10 | 20 | 30 |
|---|---|---|---|
| SSIM(%) ↑ | 81.23 ± 1.87 | 77.13 ± 2.32 | 74.14 ± 2.91 |
| PSNR(dB) ↑ | 24.60 ± 0.61 | 23.55 ± 0.62 | 21.19 ± 0.72 |

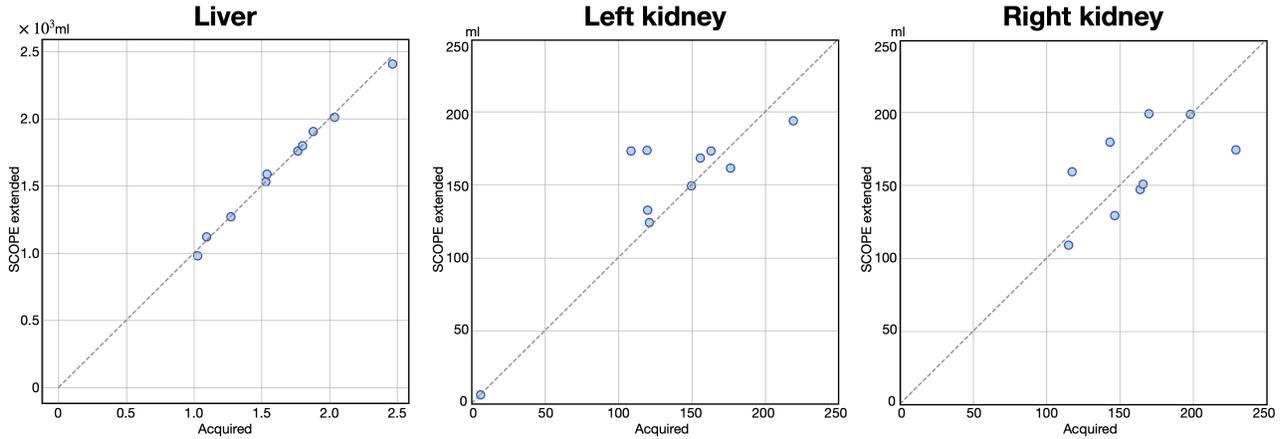

Figure 5. Volumetric agreement of the liver and kidneys between the acquired images and synthetic images.

To further evaluate SCOPE's ability in generating new slices with high anatomical fidelity, we conducted a downstream task. We used TotalSegmentator[20] to generate liver and kidney labels for both synthetic images and the acquired ground truth images, and we calculated the agreement between the two segmentation scenarios. As shown in Fig. 5, the liver shows a strong volume agreement between the synthetic and acquired images, which is expected as a significant proportion of the liver is included in the acquired FOV. This provides ample contextual information for accurate FOV extension. We define the volume disagreement ratio as

$$R_{\text{Organ}} = \frac{|V_{\text{syn}} - V_{\text{acq}}|}{V_{\text{acq}}} \times 100\% \quad (5)$$

where $V_{\text{acq}}$ and $V_{\text{syn}}$ denote the organ volume of the acquired and synthetic images, respectively. Over the 10 held-out subjects, $R_{\text{Liver}}$=1.58%±0.92%. Figure 5 (middle and right) shows the segmentation results of left and right kidney, which exhibit slightly reduced volume agreement. This can be attribute to their smaller size and partial coverage in the acquired FOV. Despite these challenges, the volume disagreement ratio over the held-out data is $R_{\text{Kidney}-L} = 11.8\% \pm 9.2\%$ for the left kidney, and $R_{\text{Kidney}-R} = 12.1\% \pm 9.8\%$ for the right kidney. An interesting observation is that there is a subject exhibiting very low volume of their left kidney. Upon detailed examination, we identified that the subject does not have a left kidney, which is a completely incidental finding.

### 3.2 Applying SCOPE on NLST

We finally applied SCOPE to $N = 100$ NLST chest CT images to extend their FOV. Since the ground truth of these extended regions is unavailable, we evaluated the results implicitly. We employed the improved BPR model[7] on SCOPE-extended images to assess whether SCOPE successfully infers and generates the missing anatomical regions. Figure 1(b) shows the results of BPR on the FOV-extended images compared to the original NLST images. The shaded area indicates that SCOPE effectively extends the FOV to include regions covering the liver and kidneys.

## 4. DISCUSSION AND CONCLUSION

In this paper, we introduced SCOPE, a novel method for extending the FOV in chest CT images using an LDM. By leveraging the natural overlapping regions in chest and abdominal CT datasets, SCOPE can generate anatomically consistent slices to cover regions beyond the initially acquired FOV. Through qualitative and quantitative evaluations, we demonstrated that SCOPE effectively extends the FOV to include critical regions such as the liver and the lungs. This standardized FOV will allow a broader application of image processing tools and motivate new findings into the human body.

Despite the promising results, our approach has some limitation that warrant future investigations. One challenge is determining where to stop the FOV extension when only specific organs, such as the liver, is needed to include. Currently, SCOPE lacks an inherent mechanism to decide the endpoint of the prediction. In future work, we aim to integrate BPR into the SCOPE pipeline to guide the synthesis process more accurately and inform the model about the anatomical "boundaries". Another limitation stems from how SCOPE relies on the natural overlapping regions to bridge chest and abdominal CT scans. Pooling datasets without care can introduce bias, such as combining a lung cancer dataset with a healthy abdominal dataset, potentially could bias the model's understanding of natural anatomical distributions. In this study, we curated our datasets to include only non-cancer cases of the NLST to mitigate this issue. Our future work will focus on expanding our method to train SCOPE on a single dataset, reducing the risk of bias and improving generalization. Additionally, while we have demonstrated that SCOPE can extend the FOV of NLST data, the broader implications of this capability remain to be explored. Future research will investigate the relationships between the lungs, liver, kidneys, and other organs within the NLST population and beyond.

In conclusion, SCOPE presents an advancement in the field of FOV extension and medical image synthesis. It demonstrates effectiveness in synthesizing extended anatomical regions from acquired CT images and enables the potential of providing deeper insights into the interplay between different organs of the human body.

## ACKNOWLEDGMENTS

**This section needs to be filled by contacting all co-authors.** This research was funded by the National Cancer Institute (NCI) grants R01 CA253923 (Landman) and R01 CA275015 (Landman).

**Declaration of the usage of generative AI:** In the creation of this work, generative AI technologies have been employed with a focused and specific purpose: to assist in structuring sentences and performing grammatical checks. It is imperative to highlight that the conceptualization, ideation, and all prompts provided to the AI originate entirely from the author(s)'s creative and intellectual efforts. The AI's function in this process has been akin to that of a sophisticated, programmatically-driven editorial tool, without any involvement in the generation of original ideas or content.